\documentclass{llncs}
\usepackage{epsf}
\usepackage[square,sort,comma,numbers]{natbib}
\usepackage{subfigure} 
\usepackage{graphicx}

\begin{document}

\renewcommand{\bibname}{References}

\title{Multi-Head Multi-Layer Attention to Deep Language Representations for Grammatical Error Detection}

\author{
Masahiro Kaneko and Mamoru Komachi}
\institute{Tokyo Metropolitan University\\
	  Graduate School of Systems Design\\
	  Tokyo, Japan, 191-0065\\
	  kaneko-masahiro@ed.tmu.ac.jp, komachi@tmu.ac.jp}
	  
\maketitle

\begin{abstract}
It is known that a deep neural network model pre-trained with large-scale data  greatly improves the accuracy of various tasks, especially when there are resource constraints.
However, the information needed to solve a given task can vary, and simply using the output of the final layer is not necessarily sufficient.
Moreover, to our knowledge, exploiting large language representation models to detect grammatical errors has not yet been studied.
In this work, we investigate the effect of utilizing information not only from the final layer but also from intermediate layers of a pre-trained language representation model to detect grammatical errors. 
We propose a multi-head multi-layer attention model that determines the appropriate layers in Bidirectional Encoder Representation from Transformers (BERT).
The proposed method achieved the best scores on three datasets for grammatical error detection tasks, outperforming the current state-of-the-art method by 6.0 points on FCE, 8.2 points on CoNLL14, and 12.2 points on JFLEG in terms of $\rm F_{0.5}$.
We also demonstrate that by using multi-head multi-layer attention, our model can exploit a broader range of information for each token in a sentence than a model that uses only the final layer's information.

\end{abstract}

\section{Introduction}

Neural networks are known to be best exploited when trained on large-scale data.
It has been demonstrated that utilizing language representation models pre-trained with large-scale data is effective for various tasks.
For example, recent studies have shown a significant improvement using large-scale data to train large deeper models for natural language understanding tasks \cite{N18-1202, OpenAI, BERT}.

In contrast, for grammatical error detection, several studies have adapted large-scale data by creating artificial training data from a large-scale raw corpora \cite{W17-5032, D18-1541}.
Moreover, there have been studies that have effectively used language representation models for grammatical error detection task \cite{P17-1194}.
To our knowledge, however, there are no studies that have utilized deep language representation models pre-trained with large-scale data for this task.

Moreover, deep neural networks learn different representations for each layer.
For example, \citet{P17-1080} demonstrated that in a machine translation task, the lower layers of the network learn to represent the word structure, while higher layers are more focused on word meaning.
\citet{N18-1202} showed that in learning deep contextualized word representations, constructing representations of layers corresponding to each task by a weighted sum improved the accuracy of six NLP tasks. 
\citet{D18-1179} empirically showed that lower layers are best-suited for local syntactic relationships, 
that higher layers better model longer-range relationships, and that the top-most layers specialize at the language modeling.
For tasks that emphasize the grammatical nature, such as grammatical error detection, information from the lower layers is considered to be important alongside more expressive information in deep layers.
Therefore, we hypothesized that using information from optimal layers suitable for a given task is important.

As such, our motivation is to construct a deep grammatical error detection model that considers optimal information from each layer. 
Therefore, we propose a model that uses multi-head multi-layer attention in order to construct hidden representations from different layers suitable for grammatical error detection.

Our contributions are as follows:
\begin{enumerate}
  \item We propose a multi-head multi-layer attention model that can acquire even more suitable representations for a given task by fine-tuning a pre-trained deep language representation model with large-scale data for grammatical error detection.
  \item We show that our model is effective at acquiring hidden representations from various layers for grammatical error detection. Our analysis reveals that using multi-head multi-layer attention effectively utilizes information from various layers. We also demonstrate that our proposed model can use a wider range of information for each token in a sentence.
   \item Experimental results show that our multi-head multi-layer attention model achieves state-of-the-art results on three grammatical error detection datasets (viz., FCE, CoNLL14, and JFLEG).
\end{enumerate}

\section{Related Works}
\subsection{Grammatical Error Detection with Language Representations}
Often, in sequence labeling tasks, recent supervised neural grammatical error detection models are built upon Bi-LSTM \cite{P16-1112, I17-1005, W17-5032, P17-1194, D18-1541, rei2018jointly}. 
\citet{rei2018jointly} used token-level predictions by Bi-LSTM for self-attention to predict sentence-level labels for grammatical error detection.
However, we adopt a transformer block-based model for token-level grammatical error detection, and we build a very deep model for this task.

\citet{P17-1194} showed the effectiveness of multitask learning by coupling language modeling and grammatical error detection. 
They used an additional objective for language modeling training to learn to predict surrounding tokens for every token in a dataset. 
In contrast to previous research, we adopt information from deep language representations for grammatical error detection by multi-head multi-layer attention.

Several studies have exploited large quantities of raw data to create additional artificial data.
\citet{W17-5032} artificially generated writing errors in order to create additional resources to learn a neural sequence labeling model following \citet{P17-1194}. 
\citet{D18-1541} employed a neural machine translation system to create error-filled artificial data for grammatical error detection.
By contrast, we directly adopt a pre-trained language representation model trained with large-scale raw data.

\subsection{Using the Layer Representations}

Deep Contextualized Word Representations (ELMo) \cite{N18-1202} used large-scale data for a deep language representation model.
Their model learns task-specific weighting from all fixed hidden layers of the pre-trained  bidirectional long short-term memory (Bi-LSTM) to construct contextualized word embeddings optimized to a given task. 
In other words, ELMo learns task-specific representations exclusively in the first layer, whereas other parameters of a pre-trained model remain unchanged.
On the contrary, we construct representations suited for given tasks by fine-tuning all parameters of our pre-trained model, using multi-head multi-layer attention.
All parameters and constructed representations of our model are trained to be best-suited for the given task.

\citet{D18-1489} employed intermediate layer representations, including input embeddings, to calculate the probability distributions in order to solve a ranking problem in language generation tasks.
Similarly, we considered the information of each layer, but our motivation is to seize the optimal information from each layer suitable for a given task using a multi-head multi-layer attention.
Moreover, their model estimated probability distributions from each layer, whereas ours constructs hidden representations from each layer for the output layer.

Furthermore, there is a study that predicts information from the middle layer of each layer of the language model and learns the errors occurring owing to the model \cite{al2018character}.
The use of the information of the middle layer of \texttt{transformer\_block} is common to our research, but the information of each layer is not taken into account at the time of evaluation and is used only for learning.
Furthermore, the information on the surface layer is less useful and learning is undertaken so that the influence of the surface layer decreases as learning progresses.
In contrast, as the method uses attention, it also lets you learn which layer is utilized in the model itself.

\section{Deep Language Representations for Grammatical Error Detection}
We propose a model that applies multi-head attention to each layer (multi-head multi-layer attention, MHMLA) to fine-tune pre-trained Bidirectional Encoder Representations from Transformers (BERT) \cite{BERT}.
Architectures of BERT and MHMLA for the grammatical error detection task are illustrated in Figure \ref{fig:architectures}.
In this section, we first introduce BERT and then explain our proposed model, MHMLA.

\subsection{BERT}
\begin{figure}%
\centering
\subfigure[][BERT.]{%
\label{fig:BERT}%
\includegraphics[scale=0.55]{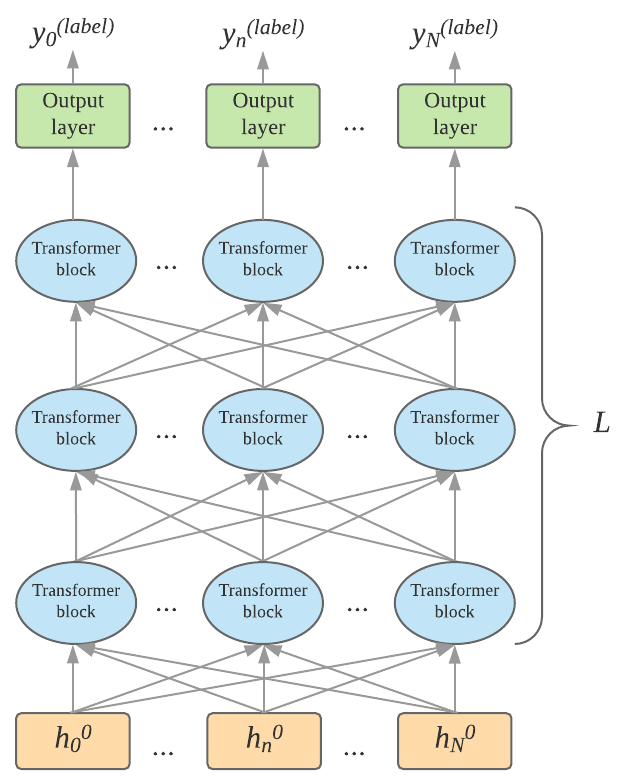} }%
\hspace{8pt}%
\subfigure[][MHMLA.]{%
\label{fig:MHMLA}%
\includegraphics[scale=0.55]{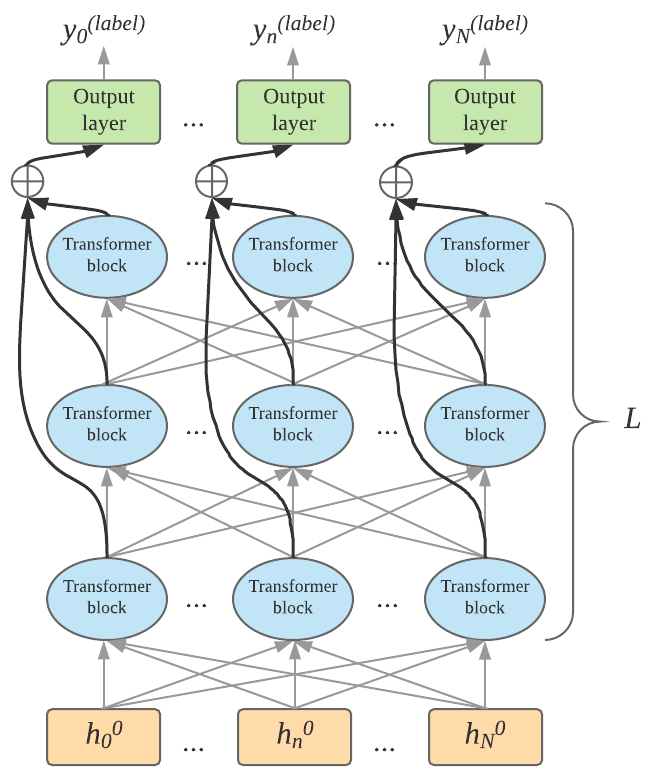}}
\caption[]{Architectures of BERT and MHMLA for grammatical error detection.}%
\label{fig:architectures}%
\end{figure}

BERT is designed to learn deep bidirectional representations by jointly conditioning both the left and right contexts in all layers (Figure \ref{fig:BERT}).
It is based on a multi-layer bidirectional transformer encoder \cite{NIPS2017_7181}. 
Insofar it is a deep language representation model pre-trained on large-scale data, it can be used for fine-tuning.
It achieved state-of-the-art results for a wide range of tasks such as natural language understanding, name entity recognition, question answering, and grounded commonsense inference \cite{BERT}.

BERT has a multi-layer bidirectional transformer encoder and can be used for different architectures, such as in classification and sequence-to-sequence learning tasks.
Here, we explain the BERT's architecture for sequence labeling tasks.
Given a sequence $S = {w_0, \cdots, w_n,\cdots, w_N}$ as input, BERT is formulated as follows:
\begin{eqnarray}
  h^0_n & = & W_{\rm e} w_n + W_{\rm p} \label{input_emb} \\
  h^l_n & = & {\rm transformer\_block}(h^{l-1}_n) \label{layer} \\
  y^{\rm (BERT)}_n & = & {\rm softmax}(W_{\rm o} h^L_n + b_{\rm o}) \label{output_layer} \label{output_bert}
\end{eqnarray}
Where $w_n$ is a current token, and $N$ denotes the sequence length.
Equation \ref{input_emb} thus creates an input embedding.
Here, \texttt{transformer\_block} includes self-attention and fully connected layers \cite{NIPS2017_7181}, and outputs $h^l_n$.
$l$ is the number of the current layer, $l \geq 1$.
$L$ is the total number of layers of BERT.
Equation \ref{output_layer} denotes the output layer.
$W_{\rm o}$ is an output weight matrix, $b_{\rm o}$ is a bias for the output layer, and $y^{\rm (BERT)}_n$ is a prediction.

The parameters $W_{\rm e}$, $W_{\rm p}$ and \texttt{transformer\_block} are pre-trained on a large document-level corpus using a masked language model \cite{taylor1953cloze} and predicting a next sentence.
Then, BERT uses a different task-specific matrix $W_{\rm o}$ of the output layer (Equation \ref{output_bert}) for a given sequence labeling task.
To adapt BERT for specific tasks, all parameters of BERT are fine-tuned jointly by predicting a task-specific label with the task-specific output layer to maximize the log-probability of the correct label.

\subsection{Multi-Head Multi-Layer Attention to Acquire Task-Specific Representations}

Multi-head attention \cite{NIPS2017_7181} is more beneficial than a single attention function.
MHMLA on a sequence labeling model applies attention to each layer $l$ of the output of \texttt{transformer\_block} $h^l_n$ of Equation \ref{layer} (Figure \ref{fig:MHMLA}).
First, we calculate attention value $v^l_n$:
\begin{eqnarray}
  v^l_{n, j} = W^l_{{\rm v} j} h^l_n + b^l_{{\rm v} j} \label{attention-start} 
 \end{eqnarray}
Here, $W_{\rm v}$ is a weight matrix, $b_{\rm v}$ is a bias, and $j$ is a head number.
We apply a non-linear layer to $h^l_n$ to acquire $k^l_n$. 
Attention score $a^l_{n}$ is as follows:
\begin{eqnarray}
  k^l_{n, j} = {\rm relu}(W^l_{{\rm k}j} h^l_n + b^l_{{\rm k}j})\\
  a^l_{n, j} = W^l_{{\rm a}j} k^l_n + b^l_{{\rm a}j}
\end{eqnarray}
where $W_{\rm k}$ and $W_{\rm a}$ are weight matrices, and $b_{\rm k}$ and $b_{\rm a}$ are biases.
Multi-heads are then calculated as follows:
\begin{eqnarray}
\tilde{a}^l_{n, j} = \frac{{\rm exp}(a^l_{n, j})}{\sum_{t=1}^L {\rm exp}(a^t_{n, j})} \label{eq:selfattention} \\
{\rm head}_{n, j} = \sum_{t=1}^L \tilde{a}^t_{n, j} v^t_{n, j} \label{attention-end}
\end{eqnarray}
where $\tilde{a}^l$ is the attention weight, normalized to sum up to 1 over all values in the layers.
These weights are then used to combine the context-conditioned hidden representations from Equation (5) into a single-token representation $c_n$:
\begin{eqnarray}
c_n = {\rm concat}({\rm head}_{n, 1}, \cdots, {\rm head}_{n, J})
\end{eqnarray}
where $J$ is the total number of heads.  
Finally, we return task-specific predictions based on this representation:
\begin{equation}
y^{\rm (label)}_n = {\rm softmax}(W_{\rm o} c_n + b_{\rm o})
\end{equation}
$W_{\rm o}$ is an output weight matrix and $b_{\rm o}$ is a bias of output layer.
Our model is optimized by minimizing cross-entropy loss on the token-level annotation.

\section{Experiments}
\subsection{Datasets}

\begin{table}[t]
\centering
\begin{tabular}{lrrr}
\hline
corpus & train & dev & test \\
\hline
FCE & 28,731 & 2,222 & 2,720 \\
CoNLL14 & - & - & 1,312  \\
JFLEG & - & - & 747  \\
\hline
\end{tabular}
\caption{Sentence statistics of used corpora.}
\label{corpora}
\end{table}

We focus on a supervised sequence labeling task: viz., grammatical error detection.
Grammatical error detection is the task of identifying incorrect tokens that need to be edited in order to produce a grammatically correct sentence.
We evaluated our approach on the three different grammatical error detection datasets.
Table \ref{corpora} shows statistics for each corpus.

\begin{description}
	\item[FCE.] We fine-tuned and searched the parameters of the model and evaluated our system on the First Certificate in English (FCE) dataset \cite{P11-1019}, which contains error-annotated short essays written by language learners.
The FCE dataset is a popular English learner corpus for grammatical error detection.
We followed the official split of the data.
	\item[CoNLL14.] We additionally used dataset from the CoNLL 2014 shared task (CoNLL14) dataset \cite{W14-1701} in our evaluation.
This dataset was written by higher-proficiency learners on different technical topics.
It was manually corrected by two separate annotators, and we report results on each of these annotations (CoNLL14-\{1,2\}).
	\item[JFLEG.] We also evaluated our approach with the JHU FLuency-Extended GUG (JFLEG) corpus \cite{E17-2037}. 
It contains a broad range of language-proficiency levels and focuses more on fluency edits and making the text more native-sounding, in addition to grammatical corrections. 
JFLEG is not labeled for grammatical error detection.
Therefore, we used dynamic programming to label tokens in sentences as correct or incorrect.
Because JFLEG is a recently developed corpus, there is only one prior study with experimental results \cite{rei2018jointly}.
JFLEG is tagged by multiple annotators, like CoNLL14, so we followed  \citet{rei2018jointly} to build a version that combines the references: if a token is labeled as an error by any annotator, it is marked as an error\footnote{Although JFLEG's experimental settings are not described in the paper, we confirmed them with the authors of the paper over e-mail.}.
\end{description}

\subsection{Experimental Details} \label{settings}
We used a publicly available pre-trained deep language representation model, namely the ${\rm BERT_{BASE}}$ uncased model\footnote{\url{https://github.com/google-research/bert}}.
This model has 12 layers, 768 hidden size, and 16 heads of self-attention.
Layer attention has 12 heads (J = 12).
We fine-tuned the model over 5 epochs with a batch size of 32.
The maximum training sentence length was 128 tokens.
We used the Adam optimizer \cite{adam} with a learning rate of 5e-05.
We applied dropout \cite{dropout} to $h^l_n$, $k^l_{n, j}$, and $\tilde{a}^l_{n, j}$ with a dropout rate of 0.3.
$\tilde{a}^l_{n, j}$ is referred to as attention dropout.
We also used WordPiece embeddings \cite{wu2016google}.
To make this compatible with sub-token tokenization, we inputted each tokenized word into the WordPiece tokenizer and used the hidden state corresponding to the first sub-token as input to the output layer, as with the original BERT.

We used ${\rm F}_{0.5}$ as the main evaluation measure.
This measure was also adopted in the CoNLL14 shared task for the grammatical error correction task \cite{W14-1701}.
It combines both precision and recall, while assigning twice as much weight to precision, because accurate feedback is often more important than coverage in error detection applications \cite{C10-2103}.

\subsection{Baselines and Comparisons}

We compare with models of \citet{P17-1194}, \citet{rei2018jointly}, \citet{W17-5032}, and \citet{D18-1541} which are based on the Bi-LSTM architecture.
The first group, \citet{P17-1194} and \citet{rei2018jointly}, was trained exclusively on the FCE dataset.
The second group, \citet{W17-5032} and \citet{D18-1541} used additional artificial data along with the FCE dataset for training. 

Our baseline and proposed models were trained on the transformer architecture. The first three are the descriptions of our baselines, and the fourth is a description of the proposed model:
\begin{description}
	\item[${\rm \bf BERT_{BASE}}$ w/o pre-train.] This model is trained using only the FCE dataset and with random initialization. This baseline did not use any other corpus for training.
	\item[${\rm \bf BERT_{BASE}}$.] This is the original pre-trained model described in Section \ref{settings} fine-tuned on the FCE dataset. This baseline uses original BERT model \cite{BERT} and can be seen as surrogated version of the proposed method without multi-layer attention.
	\item[AvgL.] This model is called averaged layers, which averages representations after linear transformation of $h^l_n$ (Equation \ref{layer}) for the output layer of ${\rm BERT_{BASE}}$ model instead of using an attention.
	\item[MHMLA.] This is the proposed model that is an extension of ${\rm BERT_{BASE}}$, with MHMLA to the pre-trained model while fine-tuning on the FCE dataset.
\end{description}

\section{Results}
 \begin{table*}[t]
 \centering
 \resizebox{1.0\textwidth}{!}{
 \begin{tabular}{l|ccc|ccc|ccc|ccc}
 \hline
  & \multicolumn{3}{|c|}{FCE} & \multicolumn{3}{|c|}{CoNLL14-1} & \multicolumn{3}{|c|}{CoNLL14-2} & \multicolumn{3}{|c}{JFLEG} \\
  & \bf P & \bf R & $\bf F_{0.5}$ & \bf P & \bf R & $\bf F_{0.5}$ & \bf P & \bf R & $\bf F_{0.5}$ & \bf P & \bf R & $\bf F_{0.5}$ \\
  \hline
  \citet{P17-1194} & 58.88 & 28.92 & 48.48 & 17.68 & 19.07 & 17.86 & 25.22 & 19.25 & 23.62 & - & - & - \\
  \citet{rei2018jointly} & 65.53 & 28.61 & 52.07 & 25.14 & 15.22 & 22.14 & 37.72 & 16.19 & 29.65 & 72.53 & 25.04 & 52.52\\
  \hline
  \citet{W17-5032} & 60.67 & 28.08 & 49.11 & 23.28 & 18.01 & 21.87 & 35.28 & 19.42 & 30.13 & - & - & - \\
  \citet{D18-1541} & - & - & 55.6 \hspace{0.95mm} & - & - & 28.3 \hspace{0.95mm} & - & - & 35.5 \hspace{0.95mm} & - & - & - \\
  \hline
   ${\rm BERT_{BASE}}$ w/o pre-train & 48.85 & 11.30 & 29.34 & 11.45 & 7.80 & 10.47 & 18.24 & 9.31 & 15.30 & 58.85 & 13.22 & 34.81 \\
  ${\rm BERT_{BASE}}$ & \bf 69.80 & 37.37 & 59.47 & 34.08 & \bf 33.56 & 33.97 & 46.01 & 33.89 & 42.93 & \bf 78.06 & 36.28 & 63.45 \\
  AvgL 		&  68.09 & 41.14 & 60.20 & 34.97 & 32.02 &34.33 &45.33& 35.27&42.88 &77.35 &37.05 & 63.52  \\
  MHMLA & $68.87^{\dagger}$ & $\bf 43.45^{\ast \dagger}$ & $\bf 61.65^{\ast \dagger}$ & $\bf 35.74^{\ast}$ & $33.50^{\dagger}$ & $\bf 35.26^{\ast \dagger}$ & $\bf 46.45^{\dagger}$ & $\bf 35.47^{\ast}$ & $\bf 43.74^{\dagger}$ & 77.74 & $\bf 38.85^{\ast \dagger}$ & $\bf 64.77^{\ast \dagger}$ \\
 \hline
 \end{tabular}
 }
  \caption{Results of grammatical error detection. These results are averaged over five runs. $\ast$ and $\dagger$ indicate that there is a significant difference against ${\rm BERT_{BASE}}$ and AvgL, respectively.} 
   \label{ged_result}
\end{table*}

 \begin{table}[t]
 \centering
 \resizebox{0.48\textwidth}{!}{
 \begin{tabular}{r|cccc}
  \hline
  $J$ & FCE & CoNLL14-1 & CoNLL14-2 & JFLEG \\
  \hline
  1 & 61.16 & 33.75 & 42.89 & 63.98 \\
  2 & 61.62 & 33.44 & 42.42 & 63.72 \\
  3 & \bf 61.90 & 34.50 & 43.17 & 64.45 \\
  4 & 61.55 & 33.74 & 42.80 & 64.37 \\
  6 & 61.22 & 34.26 & 43.29 & 64.48 \\
  8 & 61.27 & 34.72 & 43.02 & 64.10 \\
  12 & 61.65 & \bf 35.26 & \bf 43.74 & \bf 64.77 \\
   \hline
  \end{tabular}
 }
  \caption{${\rm F}_{0.5}$ scores of MHMLA using different number of heads $J$. These results are averaged over five runs. }
   \label{each_layers_results}
\end{table}

Table \ref{ged_result} shows the grammatical error detection results for the FCE, CoNLL14-\{1,2\}, and JFLEG datasets.
Scores for \citet{P17-1194}, \citet{rei2018jointly}, \citet{W17-5032}, and \citet{D18-1541} were taken from their respective papers.
In FCE, CoNLL14, and JFLEG, the ${\rm BERT_{BASE}}$ model significantly outperformed existing methods and our baseline (without pre-training) in terms of precision, recall, and $\rm F_{0.5}$.
This demonstrates that using a pre-trained deep language representation model is highly effective for grammatical error detection.
Furthermore, MHMLA achieved the highest $\rm F_{0.5}$ on all datasets, outperforming ${\rm BERT_{BASE}}$ by 2.18 points, 1.29 points, 0.81 points, and 1.32 points on FCE, CoNLL14-\{1,2\}, and JFLEG, respectively.
The scores for the AvgL model were lower than that for our proposed MHMLA model, meaning that naively using information from layers is not as effective as using MHMLA.
These results show that using MHMLA and learning task-specific representations improves the accuracy. 

\begin{figure}%
\centering
\subfigure[][FCE.]{%
\label{fig:fce}%
\includegraphics[scale=0.35]{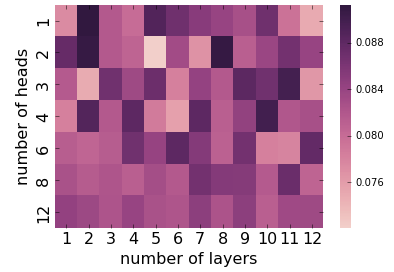}}%
\hspace{1pt}%
\subfigure[][CoNLL14.]{%
\label{fig:conll}%
\includegraphics[scale=0.35]{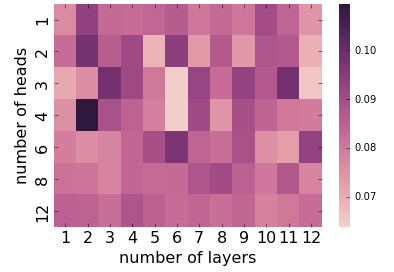}}
\hspace{1pt}%
\subfigure[][JFLEG.]{%
\label{fig:jfleg}%
\includegraphics[scale=0.35]{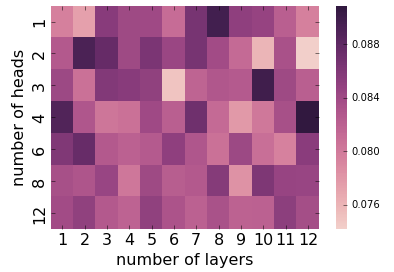}}
\caption[]{Attention visualization of MHMLA on each dataset using a different number of heads. MHMLA with 8 and 12 heads tends to attend to all layers more or less equally for all datasets.}%
\label{fig:attention}%
\end{figure}

To verify the effect of MHMLA, we examined the $\rm F_{0.5}$ value for each head number.
We investigated 1, 2, 3, 4, 6, 8, and 12 heads (i.e. the number of heads up to 12 by which the hidden layer size of 768 can be divided).
Table \ref{each_layers_results} shows the $\rm F_{0.5}$ values for each number of heads on FCE, CoNLL14-\{1,2\}, and JFLEG datasets.
Regarding FCE, the highest $\rm F_{0.5}$ score was achieved with 3 heads.
For CoNLL14-\{1,2\} and JFLEG, the $\rm F_{0.5}$ values were highest with 12 heads, demonstrating that adopting multi-head leads to improved accuracy.

\section{Analysis of the Effect of MHMLA}
The purpose of MHMLA is to construct representations not only from the final layer but also from various layers.
Multi-head attention allows the model to jointly attend to information from different representation subspaces at different positions.
Therefore, it is considered that increasing the number of heads leads to utilization of information from various layers.
Hence, we investigate the effect of the number of heads on each layer by visualizing the averaged score of MHMLA that was calculated by considering the heads $j$ of Equation \ref{eq:selfattention} for all layers on test sets of the three datasets: FCE, CoNLL14, and JFLEG.

Figure \ref{fig:attention} visualizes the average attention score to each layer of MHMLA for each head.
The average attention score is calculated by averaging ${\rm head}_{n}$ in Equation (\ref{attention-end}).
For all datasets, when there were a fewer numbers of heads, the multi-head attentions learned to attend to different layers but tended to focus on particular layers. 
For example, as shown in Figure \ref{fig:conll}, multi-head attention with heads of 2, 3, and 4 heads focused more on layers 2 and 3 while hardly attending to layers 5 and 6.
Figure \ref{fig:conll} shows that the same amount of attention is attended to each layer when the number of heads are 8 and 12.
In Figure \ref{fig:jfleg}, attention is sharp, especially with the number of heads being 1, 2, 3, and 4.
In contrast, with there are more heads, viz. 8 and 12, attention tended to attend to all layers more or less equally for all datasets.
From this visualization, we conclude that our goal of utilizing the information from various layers has been achieved. 

\section{Conclusion}

In this study, we investigated the effect of utilizing a deep language representation model (${\rm BERT_{BASE}}$) pre-trained on large-scale data for grammatical error detection.
Simply fine-tuning our ${\rm BERT_{BASE}}$ model greatly improved $\rm F_{0.5}$ scores for grammatical error detection task.

Furthermore, we have introduced an approach to learning representations suited for grammatical error detection task from various layers of a pre-trained deep language representation model using MHMLA.
Our MHMLA model outperformed previous models for grammatical error detection, establishing new state-of-the-art $\rm F_{0.5}$ scores.
Our analysis demonstrated that we succeeded at learning appropriate representations for a given task using information from different layers.

Future work includes applying MHMLA to other language representation models like Open AI GPT model \cite{OpenAI}.
Furthermore, with different combination of existing pre-trained language representation models, we hope to obtain even greater improvements.
In addition, we will explore whether our layers learned the same syntactic and semantic roles as a previous work \cite{D18-1179}, also what exactly self-attention learns at a token-level for grammatical error detection.

\section*{Acknowledgement}

This work was partially supported by JSPS Grant-in-Aid for Young Scientists (B) Grant Number JP16K16117.

\bibliography{paper}
\bibliographystyle{splncsnat}

\end{document}